# CMA-ES with Radial Basis Function Surrogate for Black-Box Optimization

Farshid Farhadi Khouzani, Abdolreza
Mirzaei, Paul La Plante and Laxmi
Gewali

**Abstract**
Evolutionary optimization algorithms often face defects and limitations that complicate the evolution processes or even prevent them from reaching the global optimum. A notable constraint pertains to the considerable quantity of function evaluations required to achieve the intended solution. This concern assumes heightened significance when addressing costly optimization problems. However, recent research has shown that integrating machine learning methods, specifically surrogate models, with evolutionary optimization can enhance various aspects of these algorithms.

Among the evolutionary algorithms, the Covariance Matrix Adaptation Evolutionary Strategy (CMA-ES) is particularly favored. This preference is due to its use of Gaussian distribution for calculating evolution and its ability to adapt optimization parameters, which reduces the need for user intervention in adjusting initial parameters.

In this research endeavor, we propose the adoption of surrogate models within the CMA-ES framework called CMA-SAO to develop an initial surrogate model that facilitates the adaptation of optimization parameters through the acquisition of pertinent information derived from the associated surrogate model. Empirical validation reveals that CMA-SAO algorithm markedly diminishes the number of function evaluations in comparison to prevailing algorithms, thereby providing a significant enhancement in operational efficiency.

## 1 Introduction

Evolutionary algorithms are commonly used for optimization because they guide solutions to better positions through mutation and selection, adapting iteratively to find optimal outcomes. Despite their power, these algorithms can struggle when target functions or constraints are computationally expensive, as each evaluation may take significant time. In such cases, the computational load can extend processing times to hours or even days, making these algorithms less feasible for complex problems. To overcome this, researchers have integrated surrogate models—approximations of target functions—that reduce the need for direct, costly evaluations, allowing faster convergence and more efficient optimization.

Among evolutionary strategies, the Covariance Matrix Adaptation Evolutionary Strategy (CMA-ES) stands out in black-box optimization for its use of Gaussian distributions to sample and search the solution space. CMA-ES iteratively adjusts distribution parameters based on previous samples, refining the solution direction effectively. However, like other evolutionary algorithms, CMA-ES demands numerous function evaluations to reach optimality, which can be impractical in high-cost problems. Integrating surrogate models with CMA-ES can help minimize these evaluations, improving efficiency without compromising solution quality.

Several surrogate modeling approaches have been applied to enhance evolutionary algorithms:

**Fitness Inheritance:** This model allows individuals to inherit fitness scores from previous evaluations, reducing the need for repeated assessments. Introduced as a local model, fitness inheritance has evolved with variations, such as three types of inherited surrogate models to further streamline the process [1].

**Gaussian Process Surrogates:** These are global models that use dimensionality reduction to manage high-dimensional optimization problems [2]. However, due to the complexity of creating accurate models across dimensions, interest has grown in local models for their adaptability.

**Neural Network Surrogates:** The first surrogate model for evolutionary strategies used a neural network to evaluate structural optimization problems, retraining periodically with new samples around the solution mean [3].

**nlmm-CMA-ES and sACM-ES Algorithms:** nlmm-CMA-ES employs a quadratic local meta-model, showing 2-4x speedups for problems with quadratic landscapes. The sACM-ES algorithm, meanwhile, uses a Ranking SVM surrogate and introduces a life-length parameter to prevent premature convergence, marking a significant innovation [4].

**Elite-driven Surrogate-assisted CMA-ES:** it combines Gaussian Process surrogates with an adaptive ILCB method and chaotic operators, enhancing diversity and efficiency, achieving up to 7.7% performance gains in complex optimization tasks [5].

**Multi Objective Surrogate-assisted KE-VFS-CMA-ES:** it integrates a Knowledge-Extraction-based Variable-Fidelity Surrogate model and CMA-ES, leveraging low-fidelity data for exploration and high-fidelity data for exploitation. With an MHVI criterion and pre-screening strategy, it enhances convergence, diversity, and efficiency, excelling in benchmark tests and aerospace engineering optimization tasks [6].

This manuscript presents a new surrogate-assisted CMA-ES approach, enhancing evolutionary optimization by leveraging surrogate-derived data to guide the search process toward high-quality solutions. The paper begins with an overview of CMA-ES principles, explains the design of the new algorithm, and provides experimental results demonstrating its efficiency, concluding with a comprehensive evaluation of the algorithm's potential to advance surrogate-assisted optimization.

## 3  CMA-ES

CMA-ES represents a repetitive stochastic optimization methodology [7], wherein each iteration entails the sampling of a population comprising candidate solutions. In contrast to traditional population-based stochastic algorithms such as Genetic Algorithms (GAs), which delineate each step as the application of operators like mutation and crossover on the population, the CMA-ES framework is conceptualized as an evolutionary transition of probabilistic distribution, facilitating the sampling of novel individuals in every generation.

In fact, the operational protocol of CMA-ES is predicated upon the following systematic steps:
- Establish a population of λ candidate solutions.
- Compute λ candidate solutions with respect to the target function.
- Adapt the sampling distribution informed by feedback derived from the function evaluations conducted in the previous step.

## 4  Proposed method

In this segment, we present our novel algorithm along with its operational mechanics; however, prior to this exposition, it is imperative to introduce the concept of model exploitation (section (3-1)) as delineated by Loschilov in [4] and adaptive encoding (section (3-2)) as articulated by Hansen [8], given that these components are integral to our algorithmic framework.

This motivates us to define visibility aware polygonal chain approximation problem as follows.

### 4.1  Model exploitation

Loshchilove introduced a surrogate-assisted Covariance Matrix Adaptation Evolution Strategy (CMA-ES) algorithm denoted as ACM-ES in [4]. A significant limitation of ACM-ES was its deficiency in surrogate control, prompting him to propose sACM-ES, which could ensure the efficacy of the surrogate model. This was achieved through the introduction of the lifelength parameter $\hat{n}$, which is adjusted in accordance with the error experienced by the surrogate $\hat{f}$ on the new test points evaluated on the function $f$.

He demonstrated that it is imperative to evaluate the quality of the surrogate model prior to its application. Upon the construction of the model, the quality is assessed on the $l_{test} = |\Lambda|$ points ordered subsequent to $f$, as measured in the following manner:

$$\text{Err}(\hat{f}) = \frac{2}{|\Lambda|(|\Lambda| - 1)} \sum_{i=1}^{|\Lambda-1|} \sum_{j=i+1}^{|\Lambda|} 1_{\hat{f}_{i,j}} \tag{1}$$

For comprehensive information, please refer to [4].

### 4.2  Adaptive encoding

Hansen introduced the Adaptive Encoding procedure in [8], which incrementally acquires a suitable coordinate system, demonstrating that the CMA-ES can be interpreted as a cumulative step-size algorithm executed within the modified coordinate system that is refined by the adaptive encoding procedure.

Kern et al. advanced the concept of coordinate transformation in [9], which is delineated from the existing Covariance matrix C and the prevailing mean value m:

$$x'_j = C^{-\frac{1}{2}} (x_j - m) \tag{2}$$

Loschilove evidenced in [4] that by transforming the training data into the newly established coordinate system as described by equation 2 and developing a surrogate model on these transformed training points, the surrogate model would exhibit an enhanced predictive capability regarding the target function. The author conducted an empirical investigation to substantiate this enhancement. We also replicate the same experimental procedure as depicted in figure (1).

It is evident that employing transformed data points facilitates the acquisition of more satisfactory contour representations of the surrogate model, which closely align with the contour lines of the rotated Ellipsoid function.

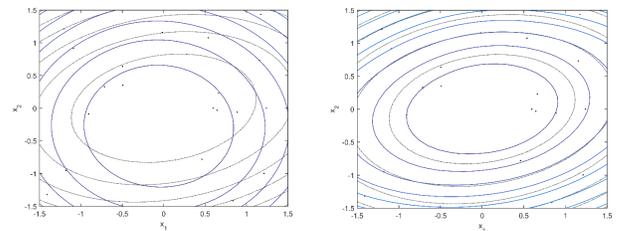

Figure 1: Contour plots of rotated Ellipsoid function (dotted lines) and cubic RBF surrogates (solid lines) obtained with standard cubic RBF model (Left) or covariance matrix-based transformed cubic RBF model (Right). The transformed cubic RBF model is more appropriate than the standard one.

### 4.3  A new surrogate-assisted optimization method

In this manuscript, we present an innovative surrogate-assisted CMA-ES framework designated as CMA-SAO

(CMA-ES surrogate-assisted optimization), which delineates a novel methodology aimed at maximizing the utilization of information derived from the surrogate models. The utilization of existing data pertaining to surrogate models in order to accelerate the search for the optimal solution represents a novel concept that has garnered significant interest among contemporary researchers. This approach has the potential to decrease the requisite number of function evaluations required to attain optimality, as it facilitates the algorithm's progression toward the optimum more expeditiously.

In the context of CMA-SAO, subsequent to the development of the surrogate model $\hat{f}$, we identify the optimum of this model $\hat{f}$ within a relatively constrained region surrounding the mean of CMA-ES, followed by the evaluation of the optimum point within the target function $f$. Should the assessed point yield a superior target function value compared to the target function value associated with the mean of CMA-ES, it will be substituted for the optimum of CMA-ES, and the covariance matrix will be adjusted based on the evolutionary trajectory from the mean to the optimum of $\hat{f}$ as delineated in Algorithm 1.

**Algorithm 1** CMA-SAO

1: given $g \leftarrow 0$; $Err \leftarrow 0.5$; $\mathcal{A}_g \leftarrow \phi$;
2: $\theta_{opt} \leftarrow$ InitializationCMA();
3: **repeat**
4: $\{\theta_{opt}, \mathcal{A}_{g+1}\} \leftarrow GenCMA(f, \theta_{opt}, \mathcal{A}_g)$;
5: $g \leftarrow g + 1$;
6: **until** $g = g_{start}$;
7: **repeat**
8: $\hat{f} \leftarrow BuildSurrogateModel(\mathcal{A}_g, \theta_{opt})$;
9: $x_{mean} = \theta_{opt}.m$;
10: $x_{best} \leftarrow FindSurrogateMinimum(\hat{f}, [x_{mean} - \frac{\xi}{2}, x_{mean} + \frac{\xi}{2}])$;
11: **if** $f(x_{best}) < f(x_{mean})$ **then**
12: $\theta_{opt} \leftarrow UpdateCMAParameters(x_{best}, \theta_{opt})$;
13: **end if**
14: **for** $i = 1, \ldots, \hat{n}$ **do**
15: $\{\theta_{opt}, \mathcal{A}_{g+1} = \mathcal{A}_g\} \leftarrow GenCMA(\hat{f}, \theta_{opt}, \mathcal{A}_g)$;
16: $g \leftarrow g + 1$;
17: **end for**
18: $\{\theta_{opt}, \mathcal{A}_{g+1}\} \leftarrow GenCMA(f, \theta_{opt}, \mathcal{A}_g)$;
19: $g \leftarrow g + 1$
20: $Err(\hat{f}) \leftarrow MeasureSurrogateError(\hat{f}, \theta_{opt})$;
21: $Err \leftarrow (1 - \beta_{Err})Err + \beta_{Err}Err(\hat{f})$;
22: $\hat{n} = \left\lfloor \frac{\tau_{err} - Err(\hat{f})}{\tau_{err}} \widehat{n_{max}} \right\rfloor$;
23: **Until** stopping criterion is met;

The CMA-SAO algorithm preserves a comprehensive global hyper-parameter vector $\theta = (\theta_{opt}, \hat{n}, \mathcal{A})$, where:
$\theta_{opt}$ denotes the optimization parameters of CMA-ES employed to ascertain the optimum of a costly target function $f(x)$.
$\hat{n}$ represents the number of generations for which the existing surrogate model is utilized.
$\mathcal{A}$ constitutes the repository of all point pairs employed for the training of the surrogate model.

In the CMA-SAO algorithm, $GenCMA(h, \theta_{opt}, \mathcal{A}_g)$ signifies one generation of CMA-ES where h is the function designated for optimization (either the actual objective $f$ or the surrogate $\hat{f}$), $\theta_{opt}$ denotes the current optimization parameters in CMA-ES (for instance, the CMA step-size and covariance matrix, among others) and $\mathcal{A}$ represents the repository of $f$.

Subsequent to each invocation of $GenCMA$, the CMA-ES parameters $\theta_{opt}$ are revised, and if the invocation was made with the true objective function, the archive $\mathcal{A}$ would consequently be updated; in other words, new points $(x, f(x))$ would be incorporated into $\mathcal{A}$.

It should also be acknowledged that prior to each construction of the surrogate model, the current points $x$ within $\mathcal{A}$ would be transformed into a new characteristic space predicated on adaptive encoding (refer to section (3-2)).

During the initial phase of the CMA-SAO algorithm, $\theta_{opt}$ and $\mathcal{A}$ are regarded as the default parameters of CMA-ES and an empty set, respectively; thereafter, to establish the training set, $GenCMA$ is executed with the target function $f$ for $g_{start}$ generations, during which all generated individuals $x$ along with their corresponding target function values $f(x)$ are incorporated into the training set $\mathcal{A}$.

After the initiation phase, the CMA-SAO engages in a repetitive six-step methodology:

Constructing a surrogate model $\hat{f}$ utilizing the existing archive $\mathcal{A}$ and the prevailing optimization parameters $\theta_{opt}$ (e.g., the matrix $C^{-\frac{1}{2}}$) to facilitate the transformation of training data into a novel characteristic space (procedure BuildSurrogateModel, line 9, section (3-3)).

Determining the minimum of the surrogate model $\hat{f}$ within a relatively constrained domain (procedure FindSurrogateMinimum, line 10) employing the function and initializing it to $x_{best}$. The constrained domain for a minimization problem is defined as follows:

$$[x_{mean} - \frac{\xi}{2}, x_{mean} + \frac{\xi}{2}] \quad (3)$$

where $x_{mean}$ denotes the current mean of the CMA-ES algorithm and $\xi$ is defined as follows:

$$\xi = 0.1 \min_{1 \leq i \leq N}(b_i - a_i) \quad (4)$$

where $N$ represents the dimensionality of the problem, $b_i$ and $a_i$ indicate the upper and lower bounds of the training set within the $ith$ dimension.

If $x_{best}$ exhibits superiority over the current mean of the CMA-ES, the CMA-ES parameters would be modified in accordance with algorithm 2 (line 11-13).

Optimizing the surrogate model $\hat{f}$ for $\hat{n}$ generations. In this phase, $GenCMA(\hat{f}, \theta_{opt}, \mathcal{A}_g)$ is consistently invoked for generations, thus $\theta_{opt}$ is revised while $\mathcal{A}$ remains static due to the absence of target function evaluations (line 14-17).

Enhancing the target function $f$ for a singular generation. During this phase, $\lambda$ individuals are assessed on $f$, the CMA-ES parameters $\theta_{opt}$ are revised, and the new points $(x, f(x))$ are incorporated into $\mathcal{A}$ (line 18).

Assessing the surrogate model error $Err(\hat{f})$ in accordance with section (3-1) based on the $\lambda$ currently evaluated points (line 20-21).

Modifying the lifelength $\hat{n}$ in accordance with section (3-1) (line 22).

**Algorithm 2** Adaptation of CMA-ES Parameters
1: $x_{mean-old} = x_{mean}$;
2: $x_{mean-new} = x_{best}$;
3: $P_\sigma \leftarrow (1 - c_{n\sigma})P_\sigma + \sqrt{c_{n\sigma}(2 - c_{n\sigma})\mu_{eff}}(C^{-\frac{1}{2}}(x_{mean-new} - x_{mean-old})/\sigma)$;
4: $P_c \leftarrow (1 - c_{cn})P_c + \sqrt{c_{cn}(2 - c_{cn})\mu_{eff}}(x_{mean-new} - x_{mean-old})/\sigma$;
5: $C \leftarrow (1 - c_{covn})C + c_{covn}P_c P_c^T$;
6: $\sigma \leftarrow \sigma \exp(\frac{c_{covn}}{d_{n\sigma}}(\frac{\|P_\sigma\|}{E\|N(0,1)\|} - 1))$;

In Algorithm 2, we modify the conjugate evolution path $P_\sigma$ by incorporating the newly generated path $(x_{mean-new} - x_{mean-old})$ into it. As referenced in [7], we initialize $\mu_{eff}$ value to 1 and delineate the learning rate of the conjugate evolution path $c_{n\sigma}$ as follows:

$$c_{n\sigma} = (\mu_{eff} + 2)/(N + \mu_{eff} + 3) \tag{5}$$

where $N$ represents the dimensionality of the problem.

The evolution path $P_c$ is adjusted utilizing the new path $(x_{mean-new} - x_{mean-old})$. The rank-one learning rate $c_{cn}$ is expressed as follows:

$$c_{cn} = (4 + \frac{\mu_{eff}}{N})/(N + 4 + 2\frac{\mu_{eff}}{N}) \tag{6}$$

Subsequent to the adaptation of the evolution path in the preceding step, the rank-one matrix $P_c P_c^T$ is integrated into the covariance matrix $C$, thereby facilitating its adaptation. The learning rate for the covariance matrix $c_{covn}$ is articulated as follows:

$$c_{covn} = \frac{2}{(N + 1.3)^2 + \mu_{eff}} \tag{7}$$

The step-size $\sigma$ is adjusted in accordance with the new conjugate evolution path. The $d_{n\sigma}$ parameter is represented as follows:

$$d_{n\sigma} = 1 + 2\max(0, \sqrt{\frac{\mu_{eff} - 1}{N + 1}} - 1) + c_{n\sigma} \tag{8}$$

### 4.4 Radial basis function model

In our methodology, we implement the Radial Basis Function (RBF) model [10] due to its strong performance compared to other surrogate models like Support Vector Machine Regression and Neural Networks, and its successful application in prior optimization techniques [11, 12]. Unlike traditional RBF learning methods used in Machine Learning, our approach tailors the model specifically for optimization. We chose a cubic RBF model, as it has demonstrated success in previous studies and is considered more suitable for surrogate-based optimization than Gaussian RBF models [12, 13].

Furthermore, we conducted a comparative analysis of the cubic RBF model against thin plate spline and linear RBF models. For the purpose of this evaluation, we executed experiments on two widely recognized functions, namely the Sphere and Rosenbrock functions, utilizing the aforementioned RBF models. To evaluate the efficacy of these models, we employed equation 1, which was introduced in section (3-1).

In our experimental framework, we constructed each model across various dimensions (2, 4, 8, 16, 32, 64, 128) utilizing a training set, and subsequently estimated the surrogate model error $Err(\hat{f})$ based on a test set, repeating this process twenty times to plot the average errors in relation to the dimensionality of the problem.

The training and test sets are composed of uniformly distributed random points within the $[-2,2]$ interval, which have been evaluated using the Sphere and Rosenbrock functions. The aggregate number of training points designated for the Sphere and Rosenbrock functions is equivalent to $2(d + 1)$ and $5(d + 1)$ respectively, with d denoting the dimensionality. The total number of test points is considered to be 10 times the total number of training points.

Figure (2) presents a graphical representation of the average evaluated errors across varying dimensions for both the Sphere and Rosenbrock functions. It is evident that for each function across all dimensions, the cubic RBF model exhibits the lowest average error when juxtaposed with the two other models.

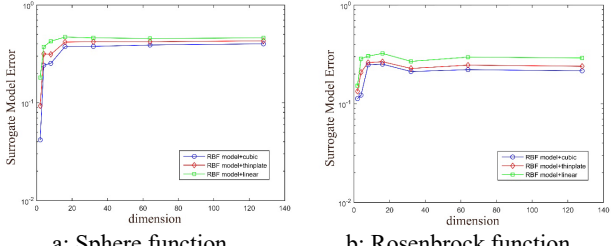

a: Sphere function    b: Rosenbrock function

Figure 2: The average errors of the surrogate model based on different functions for the RBF model concerning varying dimensions.

## 5  Experiments

Experiment 1: In the context of CMA-SAO, we commence $g_{start}$ by assigning a value of 5, which corresponds to the number of evaluations for CMA-ES during the preliminary phase, and we designate the maximum lifespan of the surrogate model $\hat{n}_{max}$ to be 20. The error threshold $\tau_{err}$ and the factor for error relaxation $\beta_{Err}$ are set at 0.45 and 0.2, respectively.

We employ the fmincon function from the Matlab optimization toolbox for the purpose of optimizing the surrogate model. Following an examination of the CMA-SAO algorithm across various dimensionalities and differing sizes of training datasets, as suggested by [7], we ascertain that the most recent data points $l$ are more advantageous for determining the number of training instances, $l$ contingent upon the dimension n as delineated below:

$$l = \lfloor 40 + 3n^{1.7} \rfloor \quad (9)$$

Initially, we compare CMA-SAO with nlmm-CMA, sACM-ES, and the conventional CMA-ES across test functions (refer to table (1)), with regard to the number of function evaluations as documented in table (2). These experiments were conducted under conditions that remain consistent for all algorithms:

- All functions considered possess an optimal value of zero.
- The results are derived from 20 independent trials. The termination criterion is the attainment of the target value $10^{-10}$, with a cap placed on the number of $1000n^2$ function evaluations.
- We evaluate all algorithms across a specified dimension [22,40].

Our tables include a column $spu$ designated as below:

$$spu = \frac{the\ number\ of\ function\ evaluations\ in\ CMA-ES}{the\ number\ of\ function\ evaluations\ in\ considered\ algorithm}$$

Table 1: Test Functions for experiments

| Name | Function | Init | $\sigma^0$ |
|---|---|---|---|
| Noisy Sphere | $f_{NoisySphere}(x) = (\sum_{i=1}^{n} x_i^2)\exp(\epsilon\mathcal{N}(0,1))$ | $[-3,7]^n$ | 5 |
| Ellipsoid | $f_{Elli}(x) = \sum_{i=1}^{n} 10^{6\frac{i-1}{n-1}} x_i^2$ | $[1,5]^n$ | 2 |
| Schwefel | $f_{Schwefel}(x) = \sum_{i=1}^{n}(\sum_{j=1}^{i} x_j)^2$ | $[-10,10]^n$ | 10 |
| Schwefel$^{1/4}$ | $f_{Schwefel^{1/4}}(x) = (f_{Schwefel}(x))^{1/4}$ | $[-10,10]^n$ | 10 |
| Rosenbrock | $f_{Rosenbrock}(x) = \sum_{i=1}^{n-1}(100(x_i^2 - x_{i+1})^2 + (x_i + 1)^2)$ | $[-5,5]^n$ | 0.5 |
| Ackley | $f_{Ackley}(x) = -20\exp(-0.2\sqrt{\frac{1}{n}\sum_{i=1}^{n} x_i^2}) + \exp(\frac{1}{n}\sum_{i=1}^{n}\cos(2\pi x_i))$ | $[1,30]^n$ | 14.5 |
| Rastrigin | $f_{Rastrigin}(x) = 10d + \sum_{i=1}^{n}(x_i^2 - 10\cos(2\pi x_i))$ | $[1,5]^n$ | 2 |
| Sphere | $f_{Sphere}(x) = \sum_{i=1}^{n} x_i^2$ | $[-10,10]^n$ | 10 |
| Cigar | $f_{Cigar}(x) = x_1^2 + 10^6 \sum_{i=2}^{n} x_i^2$ | $[-5,5]^n$ | 0.5 |
| Bohachevsky | $f_{Bohachevsky}(x) = \sum_{i=1}^{n-1}(x_i^2 + 2x_{i+1}^2 - 0.3\cos(3\pi x_i) - 0.4\cos(4\pi x_{i+1}) + .07)$ | $[1,15]^n$ | 7 |

An examination of Table (2) reveals that CMA-SAO exhibits significantly superior performance in terms of reducing the number of function evaluations when juxtaposed with two alternative algorithms, particularly in the context of high-dimensional objective functions. It is also pertinent to note that (n)lmm-CMA demonstrates enhanced efficacy exclusively on functions characterized by quadratic landscapes. Furthermore, it is essential to highlight that CMA-SAO performs commendably on multimodal functions such as $f_{Ackley}$, $f_{Rastrigin}$, and $f_{Bohachevsky}$. This observation underscores the robustness of CMA-SAO when confronted with multimodal challenges.

In Experiment 2, Loschilove conducted investigations aimed at estimating the empirical complexity associated with the training and testing of Ranking SVM, utilizing training points $\lfloor 100\sqrt{n} \rfloor$, ceasing after $\lfloor 50000\sqrt{n} \rfloor$ iterations, and evaluating the surrogate model on 500 test points (refer to Figure (3-Left)). We replicated this experiment employing identical values for both training and test points to facilitate a comparative analysis of the complexities inherent to these two surrogate models (refer to Figure (3-Right)).

For both models, the empirical complexity concerning dimension $n$ is slightly super-linear; however, it is evident that the cubic RBF model necessitates a shorter learning time in comparison to Ranking SVM. Specifically, for dimensions ranging from 100 to 200, the learning time for the cubic RBF model remains below 1 second, whereas this metric for Ranking SVM exceeds 2 seconds.

## 6  Summary and conclusion

We have introduced a novel algorithm for a surrogate-assisted Covariance Matrix Adaptation Evolution Strategy (CMA-ES), designated as CMA-SAO, with the objective of minimizing the number of function evaluations required for computationally intensive black-box optimization tasks.

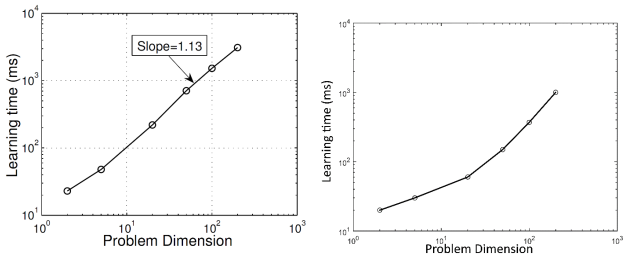

Figure 3: the cost of model learning/testing w.r.t problem dimension for Ranking SVM model (Left), cubic RBF model (Right).

The CMA-SAO algorithm employs an innovative strategy to maximize the informational yield from the surrogate model, distinguishing it from existing surrogate-assisted CMA-ES methodologies documented in the academic literature, and rendering it particularly adept for addressing high-dimensional optimization challenges. The approach adopted by CMA-SAO aims to enhance the mean position of the CMA-ES by refining the surrogate model within a relatively confined region surrounding the CMA-ES mean; subsequent to the identification of a promising point, the parameters of CMA-ES are iteratively updated to expedite the convergence toward the optimal solution of the optimization problem. Furthermore, our implementation of CMA-SAO incorporates a cubic Radial Basis Function (RBF) model. Empirical findings across ten benchmark functions demonstrate a significant advancement over the traditional CMA-ES, outperforming two additional surrogate-assisted CMA-ES algorithms, namely sACM-ES and (n)lmm-CMA.

Table 2: the mean frequency of function evaluations subsequent to twenty iterations of speedup performance (spu) for CMA-SAO, sACM-ES, (n)lmm-CMA, and CMA-ES is analyzed. The finding presented in the (n)lmm-CMA section represents the optimal values derived from the sources [5,16] (designated with a preliminary "n:" for the latter). The variable $\epsilon$ represents the intensity of noise (when applicable).

| Functio | n | λ | ε | CMA-SAO | Spu | sACM-ES | spu | (n)lmm- | spu | CMA-ES |
|---|---|---|---|---|---|---|---|---|---|---|
| $f_{Schwefel}$ | 2 | 6 | | 86 | 4.3 | 187 | 1.9 | **81** | 4.5 | 370 |
| | 4 | 8 | | **147** | 5.9 | 305 | 2.8 | **145** | 6.0 | 879 |
| | 8 | 10 | | 366 | 5.4 | 650 | 3.0 | **282** | 7.1 | 2010 |
| | 16 | 11 | | 1117 | 4.6 | 1823 | 2.8 | **627** | 8.2 | 5156 |
| | 20 | 12 | | **1635** | 4.3 | 3500 | 2.0 | | | 7042 |
| | 32 | 14 | | **3741** | 4.0 | 9689 | 1.5 | | | 15072 |
| | 40 | 15 | | **5985** | 3.7 | 16771 | 1.3 | | | 22400 |
| $f_{Schwefel^{1/4}}$ | 2 | 6 | | **398** | 3.8 | 532 | 2.8 | n:413 | 3.7 | 1527 |
| | 4 | 8 | | **835** | 3.4 | 889 | 3.2 | n:971 | 2.9 | 2847 |
| | 8 | 10 | | 1959 | 3.0 | **1852** | 3.1 | | | 5882 |
| | 16 | 11 | | **4604** | 2.6 | 4733 | 2.6 | | | 12411 |
| | 20 | 12 | | **6353** | 2.4 | 7761 | 2.0 | | | 15600 |
| | 32 | 14 | | **12349** | 2.3 | 18635 | 1.6 | | | 29378 |
| | 40 | 15 | | **20550** | 2.0 | 27365 | 1.5 | | | 41534 |
| $f_{Rosenbrock}$ | 2 | 6 | | 480 | 1.4 | 508 | 1.4 | **n:252** | 2.8 | 700 |
| | 4 | 8 | | 779 | 2.8 | 826 | 2.6 | **n:719** | 3.0 | 2187 |
| | 8 | 10 | | 2168 | 2.6 | **1802** | 3.2 | 2494 | 2.3 | 5769 |
| | 16 | 11 | | 5618 | 2.9 | **5417** | 3.0 | 7299 | 2.2 | 16317 |
| | 20 | 12 | | 7673 | 2.8 | **7218** | 3.0 | | | 21749 |
| | 32 | 14 | | **20781** | 2.5 | 24326 | 2.1 | | | 52671 |
| | 40 | 15 | | **32102** | 2.5 | 41003 | 2.0 | | | 82043 |
| $f_{NoisySphere}$ | 2 | 6 | .35 | 214 | 1.9 | 419 | - | **n:109** | 3.7 | 407 |
| | 4 | 8 | .25 | 327 | 2.5 | 864 | - | **n:236** | 3.6 | 844 |
| | 8 | 10 | .18 | **587** | 2.8 | 1232 | 1.3 | n:636 | 2.6 | 1663 |
| | 16 | 11 | .13 | **935** | 3.3 | 2089 | 1.5 | n:2156 | 1.4 | 3120 |
| | 20 | 12 | .11 | **1215** | 3.1 | 2753 | 1.4 | | | 3777 |
| | 32 | 14 | .09 | **1616** | 3.5 | 4607 | 1.2 | | | 5767 |
| | 40 | 15 | .08 | **2173** | 3.2 | 6217 | 1.1 | | | 7023 |
| $f_{Ackley}$ | 2 | 6 | | **197** | 3.7 | 269 | 2.7 | n:227 | 3.2 | 735 |
| | 4 | 8 | | **407** | 3.8 | 421 | 3.7 | | | 1577 |
| | 8 | 10 | | **689** | 4.4 | 1141 | 2.6 | | | 3066 |
| | 16 | 11 | | **1145** | 4.9 | 1451 | 3.9 | | | 5672 |
| | 20 | 12 | | **1497** | 4.4 | 3299 | 2.0 | 8150 | 0.8 | 6641 |
| | 32 | 14 | | **2521** | 3.9 | 5921 | 1.7 | | | 10063 |
| | 40 | 15 | | **3533** | 3.4 | 8456 | 1.4 | | | 12084 |
| $f_{Elli}$ | 2 | 6 | | **174** | 4.4 | 349 | 2.2 | | | 774 |
| | 4 | 8 | | **391** | 4.3 | 569 | 2.9 | | | 1688 |
| | 8 | 10 | | **1045** | 4.3 | 1121 | 4.0 | | | 4542 |
| | 16 | 11 | | **3688** | 3.5 | 2351 | 5.6 | | | 13177 |
| | 20 | 12 | | 5311 | 3.5 | **4653** | 4.0 | | | 19060 |
| | 32 | 14 | | **12015** | 3.7 | 14304 | 3.1 | | | 44562 |
| | 40 | 15 | | **19201** | 3.6 | 22101 | 3.1 | | | 69642 |
| $f_{Rastrigin}$ | 2 | 50 | | 621 | 3.1 | 1402 | 1.4 | **n:528** | 3.6 | 1970 |
| | 5 | 140 | | **3783** | 3.2 | 5121 | 2.4 | n:4037 | 3.0 | 12310 |
| | 10 | 500 | | **27105** | 1.4 | 32501 | 1.2 | 29250 | 1.3 | 40152 |
| $f_{Sphere}$ | 2 | 6 | | **81** | 4.1 | 167 | 2 | | | 334 |
| | 4 | 8 | | **129** | 5.9 | 284 | 2.7 | | | 770 |
| | 8 | 10 | | **251** | 6.2 | 589 | 2.7 | | | 1579 |
| | 16 | 11 | | **557** | 5.2 | 1196 | 2.4 | | | 2942 |
| | 20 | 12 | | **733** | 4.8 | 1790 | 2.0 | | | 3542 |
| | 32 | 14 | | **1786** | 3.0 | 3212 | 1.6 | | | 5378 |
| | 40 | 15 | | **1986** | 3.3 | 4247 | 1.5 | | | 6722 |
| $f_{Cigar}$ | 2 | 6 | | **198** | 4.1 | 274 | 2.9 | | | 818 |
| | 4 | 8 | | **373** | 4.5 | 470 | 3.6 | | | 1704 |
| | 8 | 10 | | **796** | 4.7 | 1138 | 3.3 | | | 3766 |
| | 16 | 11 | | **1511** | 4.8 | 3089 | 2.3 | | | 7343 |
| | 20 | 12 | | **1761** | 5.1 | 4262 | 2.1 | | | 8989 |
| | 32 | 14 | | **3373** | 4.0 | 8290 | 1.6 | | | 13660 |
| | 40 | 15 | | **3703** | 4.6 | 12237 | 1.3 | | | 17039 |
| $f_{Bohachevsky}$ | 2 | 6 | | **133** | 3.1 | 260 | 1.5 | | | 415 |
| | 4 | 8 | | **415** | 3.1 | 1082 | 1.2 | | | 1301 |
| | 8 | 10 | | **675** | 5.1 | 2192 | 1.5 | | | 3451 |
| | 16 | 11 | | **1450** | 6.2 | 4251 | 2.1 | | | 9001 |
| | 20 | 12 | | **2282** | 5.5 | 6282 | 2.0 | | | 12734 |
| | 32 | 14 | | **4832** | 4.5 | 8812 | 2.4 | | | 21762 |
| | 40 | 15 | | **5383** | 5.6 | 14102 | 2.1 | | | 30302 |